\documentclass[final,3p,times,twocolumn]{elsarticle}

\usepackage{amssymb}
\usepackage{amsmath}
\usepackage{lipsum}
\usepackage{booktabs}
\usepackage{makecell}

\begin{document}

\begin{frontmatter}



\title{FA-YOLO: Research On Efficient Feature Selection YOLO Improved Algorithm Based On FMDS and AGMF Modules}


\author[1,2,4]{Yukang Huo} 
\author[1,2,4]{Mingyuan Yao} 
\author[1,2,4]{Qingbin Tian} 
\author[4]{Tonghao Wang} 
\author[5]{Ruifeng Wang}
\author[1,2,3,4,6]{Haihua Wang} 

\affiliation[1]{organization={National Innovation Center for Digital Fishery},
            city={Beijing},
            postcode={100083}, 
            state={P.R},
            country={China}}
\affiliation[2]{organization={Key Laboratory of Smart Farming Technologies for Aquatic Animal and Livestock},
            addressline={Ministry of Agriculture and Rural Affairs}, 
            city={Beijing},
            postcode={100083}, 
            state={P.R},
            country={China}}
\affiliation[3]{organization={Beijing Engineering and Technology Research Center for Internet of Things in Agriculture},
            city={Beijing},
            postcode={100083}, 
            state={P.R},
            country={China}}
\affiliation[4]{organization={College of Information and Electrical Engineering},
            addressline={China Agricultural University}, 
            city={Beijing},
            postcode={100083}, 
            state={P.R},
            country={China}}
\affiliation[5]{organization={College of Engineering},
            addressline={China Agricultural University}, 
            city={Beijing},
            postcode={100083}, 
            state={P.R},
            country={China}}
\affiliation[6]{corresponding author.(e-mail: wanghaihua@cau.edu.cn)}

\begin{abstract}
Over the past few years, the YOLO series of models has emerged as one of the dominant methodologies in the realm of object detection. Many studies have advanced these baseline models by modifying their architectures, enhancing data quality, and developing new loss functions. However, current models still exhibit deficiencies in processing feature maps, such as overlooking the fusion of cross-scale features and a static fusion approach that lacks the capability for dynamic feature adjustment. To address these issues, this paper introduces an efficient Fine-grained Multi-scale Dynamic Selection Module (FMDS Module), which applies a more effective dynamic feature selection and fusion method on fine-grained multi-scale feature maps, significantly enhancing the detection accuracy of small, medium, and large-sized targets in complex environments. Furthermore, this paper proposes an Adaptive Gated Multi-branch Focus Fusion Module (AGMF Module), which utilizes multiple parallel branches to perform complementary fusion of various features captured by the gated unit branch, FMDS Module branch, and TripletAttention branch. This approach further enhances the comprehensiveness, diversity, and integrity of feature fusion. This paper has integrated the FMDS Module, AGMF Module, into Yolov9 to develop a novel object detection model named FA-YOLO. Extensive experimental results show that under identical experimental conditions, FA-YOLO achieves an outstanding 66.1$\%$ mean Average Precision (mAP) on the PASCAL VOC 2007 dataset, representing 1.0$\%$ improvement over YOLOv9's 65.1$\%$. Additionally, the detection accuracies of FA-YOLO for small, medium, and large targets are 44.1$\%$, 54.6$\%$, and 70.8$\%$, respectively, showing improvements of 2.0$\%$, 3.1$\%$, and 0.9$\%$ compared to YOLOv9's 42.1$\%$, 51.5$\%$, and 69.9$\%$.
\end{abstract}




\end{frontmatter}



\section{Introduction}
Object detection, a fundamental computer visual task, aims to identify object categories and locate their positions. It is extensively applied in various domains, including multi-object tracking \cite{pr1,pr2}, autonomous driving \cite{pr3,pr4}, robotics \cite{pr5,pr6}, and medical image analysis \cite{pr7,pr8}. With the widespread adoption of transformers, researchers have developed a series of end-to-end object detection models using the transformer's encoder-decoder architecture, such as DETR \cite{pr9}, Conditional DETR \cite{pr10}, Deformable DETR \cite{pr11}, and DINO \cite{pr12}. Although Transformer-based detectors demonstrate remarkable detection performance, they still lag behind CNN-based models in terms of speed. Over the past years, extensive research has been conducted on CNN-based detection networks, achieving significant progress \cite{pr13,pr14,pr15,pr16,pr17,pr18,pr19}. The object detection framework has evolved from two-stage models (e.g., Faster RCNN \cite{pr18} and Mask RCNN \cite{pr20}) to one-stage models (e.g., YOLO \cite{pr13}), from anchor-based (e.g., YOLOv3 \cite{pr21} and YOLOv4 \cite{pr14}) to anchor-free (e.g., CenterNet \cite{pr22}, FCOS \cite{pr23}, and YOLOX \cite{pr15}). Researchers like Golnaz Ghiasi \cite{pr24,pr25,pr26} have explored optimal network architectures for object detection tasks through NAS-FPN, and others \cite{pr27,pr28,pr29} have investigated distillation as a method to enhance model performance. The YOLO series, based on CNN models \cite{pr13,pr30,pr21,pr14,pr19,pr31,pr15,pr32,pr33}, has garnered widespread attention in the industry due to its simple structure and balance between speed and accuracy. However, YOLO series models have significant limitations in feature selection. Firstly, they lack in capturing fine-grained features and their static feature fusion method lacks the capability for dynamic adjustment, which directly impacts the model's effectiveness in complex real-world environments. Secondly, when integrating information across layers, traditional structures like FPN \cite{pr24} fail to transfer information losslessly, hindering better information fusion in YOLO. Therefore, this paper combines fine-grained and multi-scale concepts to propose an advanced feature selection module, FMDS Module (Fine-grained Multi-scale Feature Dynamic Selection Module). By adaptively and dynamically selecting fine-grained multi-scale features, this module significantly enhances feature fusion capabilities, thereby improving detection accuracy for small, medium, and large-sized targets in complex environments. Moreover, in processing images containing rich scenes and multi-scale targets, the information transfer and transformation often result in substantial loss of crucial features, limiting the model's adaptability to complex scenarios and its ultimate detection performance. Thus, to further enhance the comprehensiveness, diversity, and integrity of feature fusion, this paper introduces an Adaptive Gated Multi-branch Focus Fusion Module, consisting of multiple branches including the FMDS Module branch, a gated unit branch, and a TripletAttention branch. This module performs complementary fusion of different features captured by these branches, aiming to improve feature fusion efficiency and enhance the expression of feature maps.
We summarize the contributions of this paper as follows:

\begin{itemize}
\item This paper provides a theoretical analysis of the YOLO series of object detection models from the perspective of feature selection, revealing limitations in feature capture and fusion, as well as significant information loss during feature transmission and transformation. Based on these findings, the FMDS Module and AGMF Module were designed, achieving excellent results.

\item The FMDS Module designed in this paper enhances feature fusion capabilities by adaptively and dynamically selecting fine-grained multi-scale features, improving detection accuracy for different-sized targets in complex environments.

\item The AGMF Module, also designed in this study, integrates branches from the FMDS Module, gated unit, and TripletAttention. Through the complementary fusion of multi-branch features, it enhances feature fusion efficiency and feature expression capability.

\item Extensive experiments demonstrate that, with the implementation of the proposed FMDS Module and AGMF Module, the FA-YOLO model exhibits significant improvements in object detection performance across all metrics on the PASCAL VOC 2007 dataset.
\end{itemize}  

\section{Related work}

\subsection{Realtime object detectors}
In current, the main real-time object detectors are the YOLO series \cite{pr14,pr34,pr15,pr35,pr36,pr37,pr38,pr39,pr13,pr30,pr21,pr40,pr19,pr41,pr42,pr43}. YOLOv1-v3 \cite{pr13,pr30,pr21} established the initial YOLO framework, featuring a single-stage detection structure composed of a backbone, neck, and head, and utilized multi-scale branches to predict objects of varying sizes, thus becoming a representative single-stage object detection model. YOLOv4 \cite{pr14} optimized the previously used Darknet backbone and introduced several enhancements such as the Mish activation function, PANet, and advanced data augmentation techniques. YOLOv5 \cite{pr31}, inheriting the YOLOv4 \cite{pr14} framework, features improved data augmentation strategies and a wider variety of model variants. YOLOX \cite{pr15} integrated Multi Positives, Anchor-free, and De Coupled Head into the model structure, setting a new paradigm for YOLO model design. YOLOv6 \cite{pr39,pr38} firstly incorporate reparameterization techniques by introducing the EfficientRep Backbone and Rep-PAN Neck. YOLOv7 \cite{pr41} focused on analyzing the impact of gradient paths on model performance, proposed the E-ELAN structure to enhance model capabilities without disrupting the existing gradient pathways. YOLOv8 \cite{pr44} built upon the strengths of previous YOLO models and integrated them effectively. YOLOv9 \cite{pr36} uses GELAN to improve the architecture and training process introduced by the proposed PGI, becoming the latest generation of top-tier real-time object detectors. Although previous models employed effective feature integration methods \cite{pr24,pr45,pr46,pr47,pr48,pr49,pr50,pr51}, they still exhibited certain limitations.

\begin{figure*}
\label{fig1}
\centering
\includegraphics[width=1.1\textwidth]{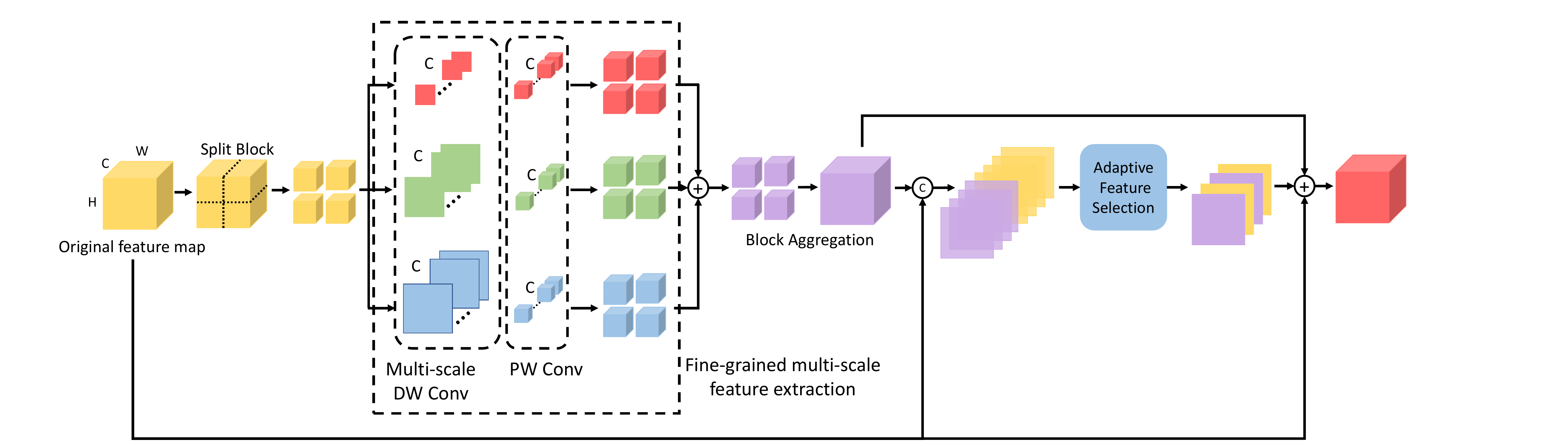}
\caption{Overall Structure Diagram of the FMDS Module}
\end{figure*}

\subsection{Multi-scale features for object detection}
Different levels of features carry positional information of different size objects. Larger feature maps contain low-dimensional texture details and the locations of smaller objects. Conversely, smaller feature maps encompass high-dimensional information and the locations of larger objects. To effectively enhance the performance of object detection, methods like SSD \cite{pr17} and DSOD \cite{pr52} added multiple convolutional layers directly after the backbone network and predict on different sized feature maps to achieve multi-scale prediction. 

Tsung-Yi Lin et al \cite{pr24}. proposed the Feature Pyramid Network (FPN), which offers an efficient architectural design by merging multi-scale features through cross-scale connections and information exchange, thereby improving the detection accuracy of objects of different sizes. DSSD \cite{pr53} utilized deconvolution for feature upsampling and accomplishes multi-scale feature fusion via element-wise multiplication. PAN et al. \cite{pr24} added a bottom-up pathway to the FPN base to further enhance information fusion, enabling high-resolution images to possess robust semantic information. EfficientDet \cite{pr47} introduced a new scalable module (BiFPN) to improve the efficiency of information fusion across different levels. Ping-Yang Chen \cite{pr54} used bidirectional fusion modules to improve interactions between deep and shallow layers. Differing from these inter-layer approaches, Chen et al. \cite{pr55} explored individual feature information using a Concentrated Feature Pyramid (CFP) method. Moreover, to address the limitations of FPN in detecting large objects, Quan et al. introduced SAFNet \cite{pr56} with adaptive feature fusion and self-enhancement modules. However, the previous FPN-based fusion structures still face issues such as slow speeds, cross-level information exchange, and information loss due to the excessive number of paths and indirect interaction methods in the network.

\subsection{Multi-branch Architectures}
The Inception architecture \cite{pr57,pr58,pr59,pr60} employed a multi-branch structure to enrich the feature space, demonstrating the importance of diversified connections, various receptive fields, and combinations of multiple branches. The Diverse Branch Block \cite{pr61} adopts the concept of using a multi-branch topology; however, it differs in that 1) the Diverse Branch Block is a building block that can be utilized across various architectures, and 2) each branch within the Diverse Branch Block can be transformed into a Conv, allowing such branches to be consolidated into a single convolution. Liu et al. \cite{pr62} input a four-channel RGB-D image into the backbone network, subsequently obtaining saliency outputs from each minor branch (single-stream network). Chen et al. \cite{pr63} utilized dual backbone networks to separately extract RGB and depth features, which were then fused using a cascaded complementary strategy (dual-stream network). Chen et al. \cite{pr64} introduced a network structure comprising two independent modal backbone networks and a parallel cross-modal distillation branch, aimed at learning complementary information. However, previous multi-branch structures rarely considered the integration of convolutional branches with attention branches, leading to the excessive weighting of redundant information. The AGMF Module, by integrating gated units, the FMDS Module, and TripletAttention, is capable of generating more comprehensive and enriched feature maps.

\begin{figure*}
\label{fig2}
\centering
\includegraphics[width=\textwidth]{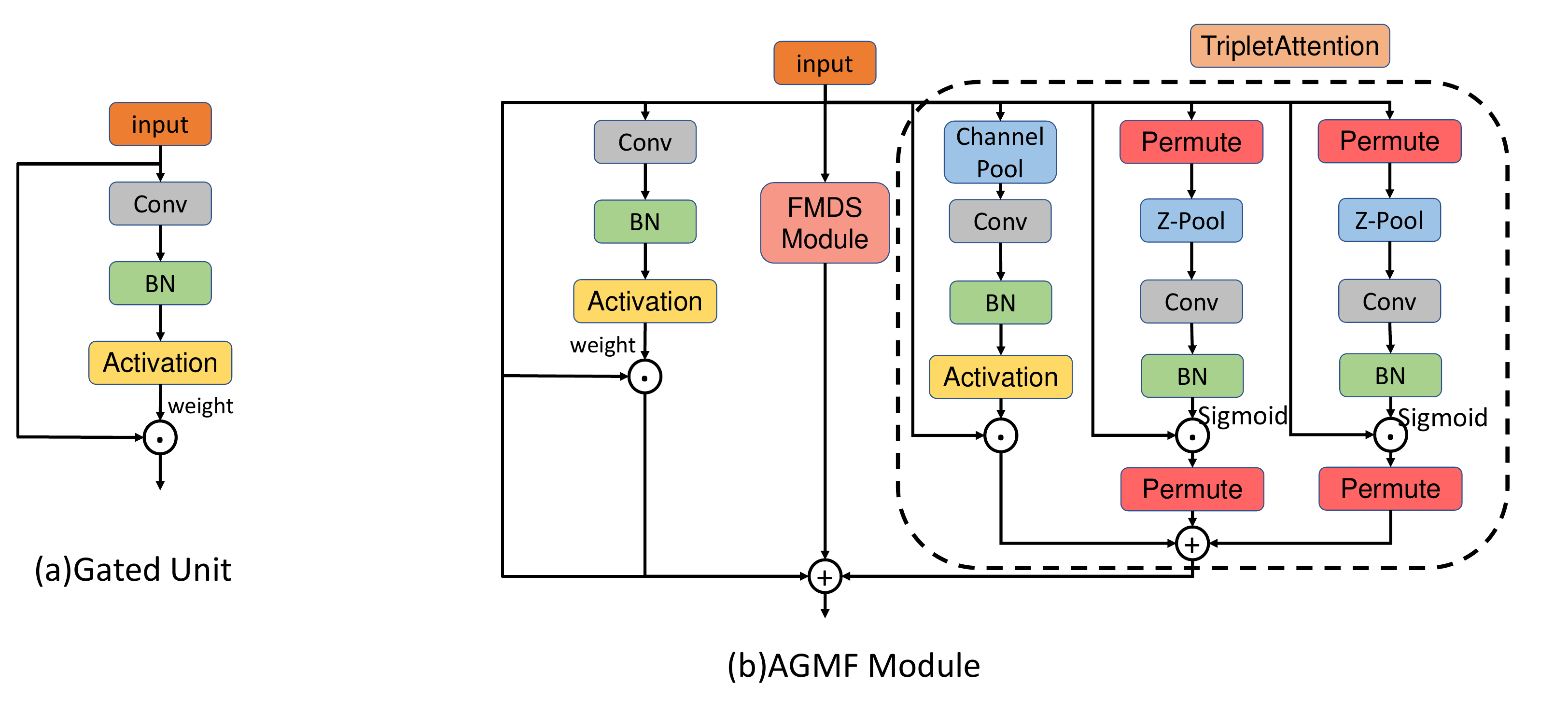}
\caption{Structure of the Gated Unit Branch and Overall Structure Diagram of the AGMF Module}
\end{figure*}

\section{Method}
\subsection{Overview}
In this section, based on the previous problem analysis, we will provide a detailed explanation of the motivations and specific structures of the FMDS Module and AGMF Module. Additionally, we will elaborate on how these two modules are integrated within the FA-YOLO framework and provide a comprehensive summary of FA-YOLO.
\subsection{FMDS Module Design}
\subsubsection{Motivation}
Although the YOLO series of models exhibit commendable detection speeds, this advantage often comes at the expense of sensitivity to detailed features. Particularly in processing small object detections or subtle changes in complex scenes, they frequently fail to capture sufficient detail information, resulting in suboptimal detection accuracy. The primary reasons are the insufficient acquisition of fine features and the lack of a feature fusion method capable of dynamically adjusting feature processing strategies according to different scenarios.

Moreover, when integrating cross-layer features, YOLO series models typically employ a structure similar to the Feature Pyramid Network (FPN) to achieve feature fusion. However, this structure encounters issues with information loss during the upward and downward transmission of information, often leading to partial loss of detail information. This loss of detail restricts the model's ability to process information and make decisions in complex environments, thereby impacting the detection performance.

\subsubsection{FMDS Module}
The FMDS Module enhances the detection accuracy of small, medium, and large-sized targets in complex environments by implementing a more efficient dynamic feature selection fusion method on fine-grained multi-scale feature maps, as illustrated in Figure 1. Initially, the FMDS Module subdivides the input feature maps into multiple smaller regional blocks, as shown in Equation \ref{eq1}, enabling the model to capture the detailed features of targets of various sizes with greater precision.

\begin{equation}
\begin{aligned}
X\_B&locks=reshape(permute(reshape
\\
&(X,(B,C,N,H\,\,// 2,2,W\,\,// 2)),
\\
&(0,2,4,1,3,5)),(-1,C,H\,\,// 2,W\,\,// 2))
\end{aligned}
\label{eq1}
\end{equation}

In this context, X represents the input feature map, B denotes the batch size, and H and W respectively stand for the height and width of the feature map. X\_Blocks signifies the subdivision of the feature map into multiple smaller regional blocks.
Subsequently, each regional block is independently processed by convolutional kernels of different scales, as illustrated in Equation \ref{eq2}. This approach not only enhances the local sensitivity of the features but also enables the model to capture more detailed spatial hierarchical information.

\begin{equation}
\begin{aligned}
X\_Blocks'=&\sum_{i=1}^3f_{{pw}}(f_{{dw}}(X\_Blocks,K_{dw\_i},S_{dw\_i},P_{dw\_i}),
\\
&K_{pw\_i},S_{pw\_i},P_{pw\_i})
\end{aligned}
\label{eq2}
\end{equation}

Within this framework, $f_{dw}$ represents the Depthwise Convolution, where $K_{dw\_i}$, $S_{dw\_i}$, and $P_{dw\_i}$ respectively denote the size, stride, and padding of the Depthwise Convolution kernel. Similarly, $f_{pw}$ denotes the Pointwise Convolution, with $K_{pw\_i}$, $S_{pw\_i}$, and $P_{pw\_i}$ specifying the size, stride, and padding of the Pointwise Convolution kernel. X\_Blocks' represents the regional blocks obtained after being processed independently by convolutional kernels of various scales.

Subsequently, the processed fine-grained multi-scale features are integrated, as shown in Equation \ref{eq3}.
\begin{equation}
\begin{split}
\begin{aligned}
X'=&reshape(permute(reshape
\\
&( X\_Blocks',( B,4,-1,H\,\,//2,W\,\,//2 ) ) ,
\\
&(0,2,1,3,4),(B,-1,H,W)
\end{aligned}
\end{split}
\label{eq3}
\end{equation}

In Equation \ref{eq3}, $X'$ results from reassembling the multiple processed regional blocks into a processed feature map. 

The integrated feature map is then concatenated with the original feature map to form a new feature map, as illustrated in Equation \ref{eq4}. 

\begin{equation}
\begin{split}
\begin{aligned}
\mathrm{  }X_{concat}=concat\!\:\left( X,X',dim=1 \right) 
\end{aligned}
\end{split}
\label{eq4}
\end{equation}

In this model, $X_{concat}$ is the feature map obtained by concatenating the original feature map with the feature map derived from Equation \ref{eq3} along the first dimension.

Subsequently, the new feature map is dynamically evaluated to determine the importance of features from different regional blocks and scales, optimizing the weight distribution of the features, as shown in Equation \ref{eq5}.

\begin{equation}
\begin{split}
\begin{aligned}
X'_{select}&=DepthwiseSeparableConv(X_{concat},
\\
&in\_channels*2,out\_channels,K,S,P)
\end{aligned}
\end{split}
\label{eq5}
\end{equation}

In this model, $X'_{select}$ refers to the feature map that has undergone feature-adaptive selection. K, S, and P respectively represent the size, stride, and padding of the convolution kernel.

This fine-grained and multi-scale dynamic feature selection and fusion approach significantly enhances the model's ability to handle small, medium, and large-sized targets in complex environments. For small-sized target detection, the model improves recognition rates by focusing on finer granularity of feature details; for medium to large-sized targets, it achieves more accurate localization and recognition by integrating multi-scale features to obtain sufficient contextual support.

\begin{figure*}
\label{fig3}
\centering
\includegraphics[width=\textwidth]{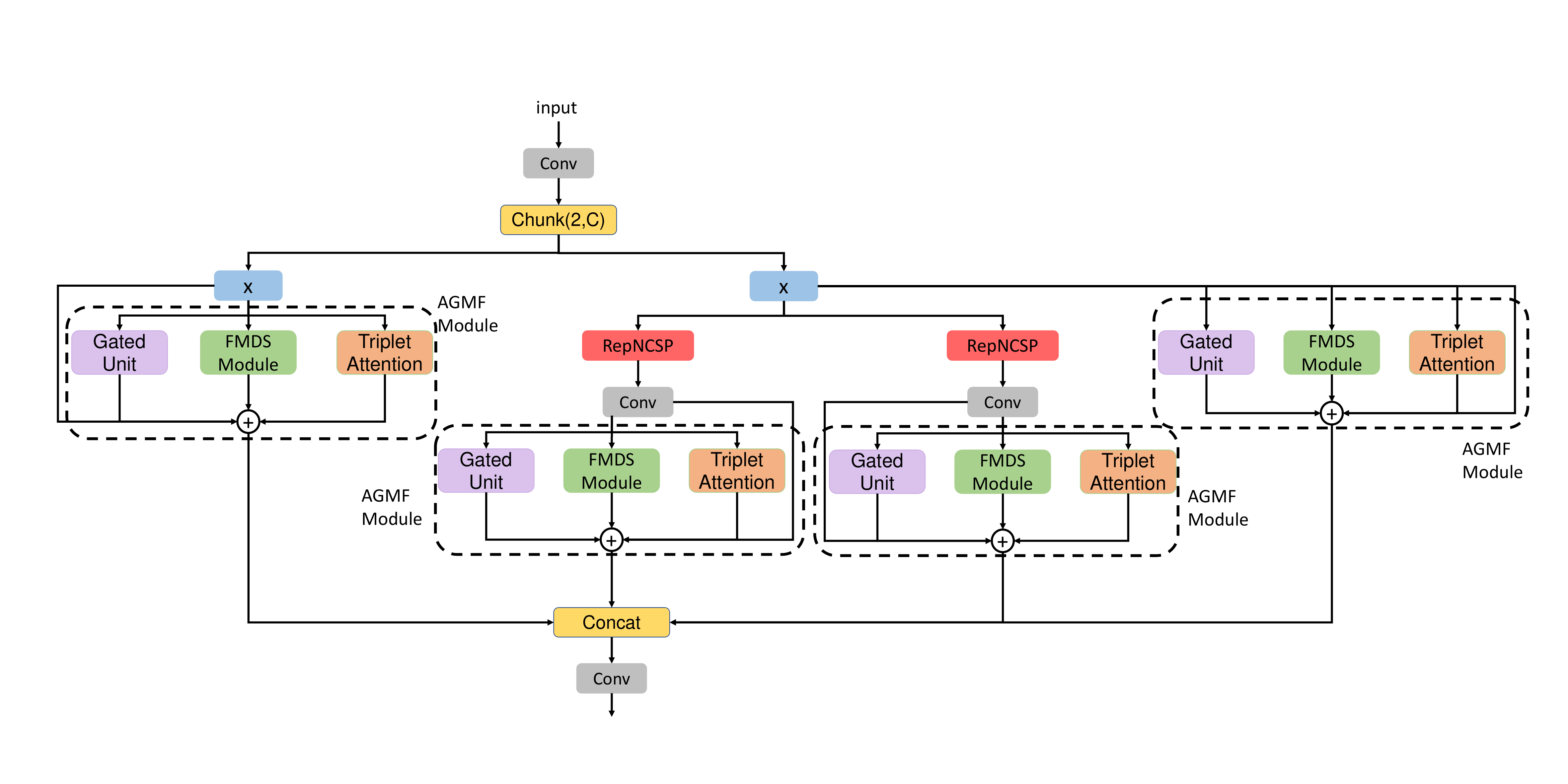}
\caption{Overall Structure Diagram After Integrating the FMDS Module and AGMF Module into the RepNCSPELAN4 Module}
\end{figure*}

\subsection{AGMF Module Design}
\subsubsection{Motivation}
Different data characteristics, such as texture, color, and semantic content, may require distinct processing strategies. A single processing branch often fails to comprehensively capture the multidimensional features of complex data, especially when these features are intertwined and interdependent. However, a single processing branch may struggle to adapt to the variations and demands of different types of data, potentially leading to the loss of important information or insufficient feature representation. While traditional multi-convolution branches possess more robust capabilities for spatial feature extraction, they too can result in the overemphasis of irrelevant information. On the other hand, attention branches focus on key information and relationships within the data, optimizing the capability to parse global contexts. The combination of the convolution branch and the attention branch can enable the model to not only capture details accurately but also grasp essential features and trends of the overall data.

\subsubsection{AGMF Module}
The AGMF module is designed with three main parallel processing branches: the Gated Unit branch, the FMDS Module branch, and the TripleAttention branch. Each branch is responsible for capturing and processing different aspects of the data. While efficiently collaborating, these branches also maintain the module's flexibility and high performance. The overall structure of the module is illustrated in Figure2(b).

The Gated Unit branch regulates and controls the flow of information in the feature maps, allowing it to adaptively adjust based on the dynamic changes in the data. It filters information critical to the current task, suppresses irrelevant or redundant data transmission, and enhances the model's focus and efficiency, as illustrated in Equations \ref{eq6} and \ref{eq7}.

\begin{equation}
\begin{split}
\begin{aligned}
Y_{i\_GU\_weight\mathrm{ }}=Activation\!\:\left( BN\!\:\left( Conv2d\left( Y_i \right) \right) \right) 
\end{aligned}
\end{split}
\label{eq6}
\end{equation}

In this configuration, $Y_{i}$ represents the input feature map, and $Y_{i\_GU\_weight}$ denotes the weight assigned by the Gated Unit.

\begin{equation}
\begin{split}
\begin{aligned}
Y_{i\_GU}=Y_i\cdot Y_{i\_GU\_weight}
\end{aligned}
\end{split}
\label{eq7}
\end{equation}

In Equation \ref{eq7},$Y_{i\_GU}$ represents the output from the Gated Unit branch.

The FMDS Module branch effectively captures and selects multi-scale and fine-grained data features, enabling it to gather a range of feature scales from details to global aspects from the input data. The TripleAttention branch utilizes the TripleAttention attention mechanism to focus on enhancing the model’s recognition and processing capabilities for key features, thereby strengthening the model's ability to identify crucial data characteristics. After each of these three branches has processed the data independently, their respective feature outputs are gathered into a fusion layer. This layer considers the importance and complementarity of each branch's outputs, integrating these features to form a final, high-quality feature representation.

\subsection{Architecture Design of FA-YOLO}
To ensure consistency in subsequent ablation experiments, FA-YOLO employs the same data augmentation strategy and hyperparameter settings as YOLOv9. The primary difference between FA-YOLO and YOLOv9 is the introduction of the FMDS Module and AGMF Module within the RepNCSPELAN4 module. These additions significantly enhance feature fusion and feature representation capabilities, improving the detection accuracy of targets of various sizes in complex environments. The structural diagram is shown in Figure 3.

\section{Experiment and Result analysis}
\subsection{Setups}
\subsubsection{Datasets}
We conducted extensive experiments using the PASCAL VOC 2007 dataset to validate the proposed FA-YOLO enhancement algorithm. All of our experiments were conducted without the use of pre-trained models; instead, all models were trained from scratch. Finally, we compared the detection performance of FA-YOLO with other mainstream models in the YOLO series on the PASCAL VOC 2007 dataset.

\begin{table*}[]
\centering
\begin{tabular}{llrrrrrrrr}
\hline
model    & Anchor      & Param (M) & FLOPs (G) & mAP  & $AP_{50}$ & $AP_{5}$ & $AP_{S}$  & $AP_{M}$  & $AP_{L}$  \\
\hline
YOLOv5-S\cite{pr31} & Anchor-Base & 7.2       & 16.5      & 43.4 & 70.1 & 47.8 & 25.0   & 36.9 & 47.2 \\
YOLOv5-M\cite{pr31} & Anchor-Base & 21.2      & 49.0        & 50.2 & 73.0   & 55.8 & 25.7 & 41.8 & 55   \\
YOLOv5-L\cite{pr31} & Anchor-Base & 46.5      & 109.1     & 51.4 & 74.6 & 57.5 & 26.7 & 43.2 & 55.4 \\
YOLOv7\cite{pr41}   & Anchor-Base & 36.9      & 104.7     & 55.4 & 77.9 & 61.7 & 24.2 & 40.4 & 61.6 \\
YOLOv8-N\cite{pr44} & Anchor-Free & 3.2       & 8.7       & 54.5 & 74.6 & 60.8 & -    & -    & -    \\
YOLOv8-S\cite{pr44} & Anchor-Free & 11.2      & 28.6      & 59.0   & 78.3 & 64.9 & -    & -    & -    \\
YOLOv8-M\cite{pr44} & Anchor-Free & 25.9      & 78.9      & 63.1 & 81.1 & 69.0   & -    & -    & -    \\
YOLOv8-L\cite{pr44} & Anchor-Free & 43.7      & 165.2     & 65.3 & 82.9 & 71.9 & -    & -    & -    \\
YOLOv9\cite{pr36}   & Anchor-Free & 25.5      & 102.8     & 65.1 & 84.5 & 72.8 & 42.1 & 51.5 & 69.9 \\ 
FA-YOLO(Ours)  & Anchor-Free & 30.7      & 101.9     & \textbf{66.1} & \textbf{85.1} & \textbf{73.1} & \textbf{44.1} & \textbf{54.6} & \textbf{70.8} \\ \hline
\end{tabular}
\caption{FA-YOLO compared to other YOLO series models}
\label{tab1}
\end{table*}

\begin{table*}[]
\centering
\begin{tabular}{llrrrrrr}
\hline
model    & input size & mAP  & $AP_{50}$ & $AP_{75}$ & FPS(bs=1) & FPS(bs=32) & latency(bs=1) \\ \hline
YOLOv5-S\cite{pr31} & 640        & 43.4 & 70.1 & 47.8 & 376.0       & 444.0        & 2.7ms         \\
YOLOv5-M\cite{pr31} & 640        & 50.2 & 73.0   & 55.8 & 182.0       & 209.0        & 5.5ms         \\
YOLOv5-L\cite{pr31} & 640        & 51.4 & 74.6 & 57.5 & 113.0       & 126.0        & 8.8ms         \\
YOLOv7\cite{pr41}   & 640        & 55.4 & 77.9 & 61.7 & 110.0       & 122.0        & 9.0ms         \\
YOLOv8-N\cite{pr44} & 640        & 54.5 & 74.6 & 60.8 & 561.0       & 734.0        & 1.8ms         \\
YOLOv8-S\cite{pr44} & 640        & 59.0   & 78.3 & 64.9 & 311.0       & 387.0        & 3.2ms         \\
YOLOv8-M\cite{pr44} & 640        & 63.1 & 81.1 & 69.0   & 143.0       & 176.0        & 7.0ms         \\
YOLOv8-L\cite{pr44} & 640        & 65.3 & 82.9 & 71.9 & 91.0        & 105.0        & 11.0ms        \\

FA-YOLO(Ours)  & 640        & \textbf{66.1} & \textbf{85.1} & \textbf{73.1} & 88        & 99.2       & 11.2ms  
\\ \hline
\end{tabular}
\caption{Comparison of FPS and Latency Between FA-YOLO and Other YOLO Series Models. The input image size for the comparison is 640}
\label{tab2}
\end{table*}

\subsubsection{Implementation details}
This paper adopts the settings of YOLOv9, using the same architecture and training configurations, with the exception of the RepNCSPELAN4 structure.The optimizer and other setting are also same as YOLOv9, i.e. stochastic gradient descent (SGD) with momentum and cosine decay on learning rate. Warm-up, grouped weight decay strategy and the exponential moving average (EMA) are utilized. The data augmentations we adopt are Mosaic and Mixup. The batch size is set as 32. The total number of training times is 500 epochs. In setting the learning rate, we use linear warm-up in the first three epochs, and the subsequent epochs set the  corresponding decay manner according to the model scale. As for the last 15 epochs, we turn mosaic data augmentation off.

\subsection{Results}
The method proposed in this paper is compared with models from the YOLO series, and the results are shown in Tables 1 and 2. The implementation results indicate that our proposed FA-YOLO significantly surpasses the existing mainstream YOLO series models in detection performance. Compared to YOLOv5-L, FA-YOLO has a 14.7\% higher mAP, and its parameter count is only 30.7 M, approximately two-thirds that of the latter. Compared to YOLOv7, FA-YOLO has 6.2 M fewer parameters, accounting for 20\% of the total parameters of the FA-YOLO model, and FA-YOLO's mAP and $AP_{50}$ are 10.7 and 7.2 points higher than those of YOLOv7, respectively. FA-YOLO has only 70.3\% of the parameters of YOLOv8-L but achieves a 0.8 point higher mAP, with almost the same throughput (batch size of 1) and GPU latency as YOLOv8-L. Compared to YOLOv9, FA-YOLO shows a 1.0\% improvement in mean Average Precision (mAP), and the accuracies for detecting small, medium, and large targets are 2.0\%, 3.1\%, and 0.9\% higher, respectively. Moreover, FA-YOLO's GFLOPs are 101.9, which is less than YOLOv9's 102.8, indicating that FA-YOLO requires lower computational costs compared to YOLOv9.

\begin{figure*}
\label{fig4}
\centering
\includegraphics[width=\textwidth]{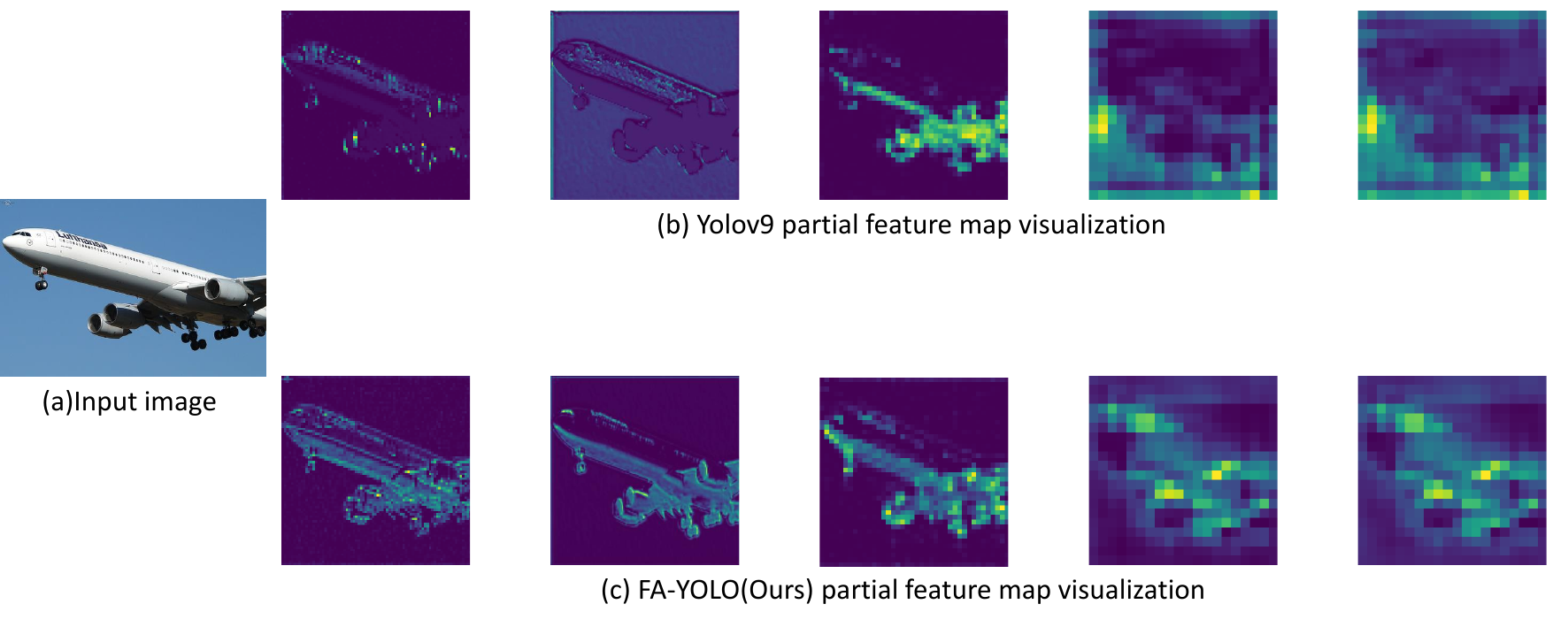}
\caption{Comparison of Visualization Results of Feature Maps at the Same Layer between FA-YOLO and YOLOv9}
\end{figure*}

\begin{table*}[!ht]
    \centering
    \begin{tabular}{|c|c|c|c|r|r|r|r|r|}
    \hline
        baseline & FMDS  & AGMF  & \thead{AGMF Module((Remove the FMDS Branch, \\ Retaining Only the Other Two Branches))}  & mAP & $AP_{50}$ & $AP_{S}$ & $AP_{M}$ & $AP_{L}$  \\ \hline
        \checkmark &  &  &  & 65.1 & 84.5 & 42.1 & 51.5 & 69.9  \\ \hline
        \checkmark & ~ & ~ & \checkmark & 65.7 & 84.8 & 42.3 & 53.0 & 70.9  \\ \hline
        \checkmark & \checkmark & ~ & ~ & 65.5 & 84.9 & 43.7 & 53.6 & 70.4  \\ \hline
        \checkmark & \checkmark & \checkmark & ~ & \textbf{66.1}  & \textbf{85.1} & \textbf{44.1} & \textbf{54.6} & \textbf{70.8}  \\ \hline
    \end{tabular}
    \caption{FA-YOLO Component Ablation Experiment}
    \label{tab3}
\end{table*}

\subsection{Ablations}
To validate the effectiveness of our feature fusion analysis and evaluate the proposed FMDS Module and AGMF Module, we independently examined each module within FA-YOLO, focusing on mAP, $AP_{50}$, $AP_{S}$, $AP_{M}$, and $AP_{L}$, as shown in Table 3. The results demonstrate that the FMDS Module, by implementing a more efficient dynamic feature selection and fusion method on fine-grained multi-scale feature maps, significantly enhances the detection accuracy of small, medium, and large-sized targets in complex environments, achieving a 0.4\% mAP performance gain. Particularly, the detection accuracies for small and medium targets have improved by 2.0\% and 3.1\%, respectively. The AGMF Module, by integrating the outputs of the FMDS Module, TripletAttention branch, and Gated Unit branch, performs a complementary fusion of the different features captured by multiple branches to form a final high-quality feature representation, achieving a 1.0\% increase in mAP. Additionally, when the FMDS Module branch is removed from the AGMF Module (leaving only the Gated Unit and TripletAttention branches), the mAP and the accuracy for small target detection decrease from 66.1\% and 44.1\% to 65.7\% and 42.3\%, respectively. Compared to before the removal of the FMDS Module, the mAP and $AP_{S}$ decrease by 0.4\% and 1.8\%, respectively.

\subsection{Visualization}
In this paper, we propose the FMDS Module, which achieves more efficient feature fusion through adaptive dynamic selection of fine-grained multi-scale features. Additionally, the AGMF Module combines the outputs of the FMDS Module, the TripleAttention branch, and the Gated Unit branch. By performing a complementary fusion of the different features captured by these multiple branches, it effectively enhances the efficiency of feature fusion and strengthens the expressive capacity of the feature maps. To validate the effectiveness of these designs, this paper employs a visualization comparison of the feature maps from the same layers of FA-YOLO and YOLOv9, as illustrated in Figure 4. Figure 4(b) shows the visualization results of the feature maps from YOLOv9, while Figure 4(c) presents the feature map visualization structure of FA-YOLO. As can be clearly seen from Figure 4, FA-YOLO, which integrates the FMDS and AGMF Modules, demonstrates significant improvements in the detection and localization capabilities within the images.

\section{Conclusion}
In this paper, we conduct an in-depth analysis of the limitations of the YOLO series models in feature capture and fusion, particularly identifying significant losses of important features during the feature transmission and transformation process. To address this issue, we designed the FMDS and AGMF Modules and validated their effectiveness through experiments.

The FMDS Module enhances feature fusion capabilities through adaptive dynamic selection of fine-grained multi-scale features, significantly improving the model's detection accuracy for various sized targets in complex environments. Additionally, the AGMF Module integrates the branches of the FMDS Module, the Gated Unit, and Triplet Attention. This integration of multiple branch features through complementary fusion further enhances feature fusion efficiency and expressive capacity.

Based on the design of the FMDS and AGMF Modules, we propose a new object detection model named FA-YOLO. Compared to the latest YOLOv9, FA-YOLO shows superior performance improvements: the mean Average Precision (mAP) increase by 1.0\%, $AP_{50}$ by 0.6\%, and AP75 by 0.3\%. Particularly in the detection accuracy of different sized targets, there is an improvement of 2.0\% in small-sized targets ($AP_{S}$), 3.1\% in medium-sized targets ($AP_{M}$), and 0.9\% in large-sized targets ($AP_{L}$). These results significantly demonstrate the efficiency and accuracy of FA-YOLO in object detection tasks.

\section{Acknowledgments}
\subsection{Funding}
Mandarin fish factory farming service project (202305510811525) and key technology research and creation of digital fishery intelligent equipment (2021TZXD006).
\subsection{Conflict of interest}
We have no conflict of interest.








\bibliographystyle{elsarticle-num}
\bibliography{bibfile}


\end{document}